\documentclass[conference]{IEEEtran}

\IEEEoverridecommandlockouts
\usepackage{cite}
\usepackage{xurl}

\usepackage{amsmath,amssymb,amsfonts}
\usepackage{algorithmic}
\usepackage{graphicx}
\usepackage{textcomp}
\usepackage{xcolor}
\def\BibTeX{{\rm B\kern-.05em{\sc i\kern-.025em b}\kern-.08em
    T\kern-.1667em\lower.7ex\hbox{E}\kern-.125emX}}
\usepackage[colorlinks,citecolor=red,urlcolor=blue,bookmarks=false,hypertexnames=true]{hyperref} 
\begin{document}

\title{Towards a Multimodal System for Precision Agriculture using IoT and Machine Learning\\
{\scriptsize \textsuperscript{*}Note: This paper is accepted in the 12th ICCCNT 2021 conference at IIT Kharagpur, India. The final version of this paper will appear in the conference proceedings.}}

%\title{Towards a Multimodal System for Precision Agriculture using IoT and Machine Learning\\
%{\scriptsize Note: Copyright 2021 IEEE. Published in the 12th International Conference on Computing, Communication and Networking Technologies (ICCCNT 2021, scheduled for July 6-8 (2021) in IIT Kharagpur, WB, India. Personal use of this material is permitted. However, permission to reprint/republish this material for advertising or promotional purposes or for creating new collective works for resale or redistribution to servers or lists, or to reuse any copyrighted component of this work in other works, must be obtained from the IEEE. Contact: Manager, Copyrights and Permissions / IEEE Service Center / 445 Hoes Lane / P.O. Box 1331 / Piscataway, NJ 08855-1331, USA.}}

\author{\IEEEauthorblockN{Satvik Garg\textsuperscript{1}, Pradyumn Pundir\textsuperscript{1},
Himanshu Jindal\textsuperscript{1}, Hemraj Saini\textsuperscript{1}, Somya Garg\textsuperscript{2}}
\IEEEauthorblockA{\textsuperscript{1}Jaypee University of Information Technology, Solan, India\\
\textsuperscript{2}Deloitte Consulting LLP, New York, USA}
Email: {\{satvikgarg27, pundirpradyumn25, himanshu19j\}@gmail.com}, hemraj1977@yahoo.co.in, somgarg@deloitte.com}
\maketitle

\begin{abstract}
Precision agriculture system is an arising idea that refers to overseeing farms utilizing current information and communication technologies to improve the quantity and quality of yields while advancing the human work required. The automation requires the assortment of information given by the sensors such as soil, water, light, humidity, temperature for additional information to furnish the operator with exact data to acquire excellent yield to farmers. In this work, a study is proposed that incorporates all common state-of-the-art approaches for precision agriculture use. Technologies like the Internet of Things (IoT) for data collection, machine Learning for crop damage prediction, and deep learning for crop disease detection is used. The data collection using IoT is responsible for the measure of moisture levels for smart irrigation, n, p, k estimations of fertilizers for best yield development. For crop damage prediction, various algorithms like Random Forest (RF), Light gradient boosting machine (LGBM), XGBoost (XGB), Decision Tree (DT) and K Nearest Neighbor (KNN) are used. Subsequently, Pre-Trained Convolutional Neural Network (CNN) models such as VGG16, Resnet50, and DenseNet121 are also trained to check if the crop was tainted with some illness or not.  
\end{abstract}


\section{Introduction}
As per the report by the National Crime Record Bureau (NCRB), the number of farmers died in 2019-2020 is approximately 10,281 [1]. Despite the popular image of farmers, suicide in agriculture has become more common. Numerous workers still adopt the regular methods in farming which brings about low yield and efficiency. A study conducted by the Centre for Study of Developing Societies (CSDS) [2] found that 76\% of the farmers want to quit farming. It also reports that 74\% of farmers didn't get basic farming-related information such as fertilizer doses from officials of the agriculture department. Hence, the precision agriculture system assists and helps the farmers with robotizing and upgrading them to improve rural profitability and contribute to making farming systems smart [3]. The introduction of new technologies such as IoT-based devices would definitely have a positive impact. For the most part, large-scale setups are hard to execute in real-world scenarios. This paper involves using existing advancements like, for instance, sensor-based modules that are demanding and easy to carry out.

An IoT-based smart irrigation system is designed which maintains a balance in the water level of the field. It turns the water pump 'on' when the field is dry and turns it 'off' when it is wet. Also, considering one of the major problems in today’s world is water scarcity [4], this system will help managing water levels and only irrigates fields when needed. %In a smart IoT-based irrigation system, a soil moisture sensor [3] is dug into the soil that gathers the information about the soil moisture content present in the soil and signals relay module whether the field is dry or wet and when to turn on/off water pump accordingly. Relay module [4] is an electrical circuit that aims to open or close a circuit. A web portal is also designed using HTML, CSS, and Javascript to monitor the crop and an individual can communicate water pump through that web portal.%

Crop mainly required three macronutrients: n (nitrogen), p (phosphorus), k (potassium). Shortage of nutrients can cause deficiency and affect crop’s health. Fertilizers are used to provide nutrients to the soil; it is a natural or chemical substance. Overuse of fertilizers can have a negative impact not only on the crops but also on the groundwater. So, a fertilizers dose recommendation system can help to increase crop productivity and controls the overuse of fertilizers. Using firebase, data can be stored with help of an IoT-based fertilizers dose recommendation system. One can easily monitor the results from the previously collected data in the firebase database. %It is a cloud-hosted database service%

Deep learning is becoming popular and common since the introduction of ImageNet by Alex Krizhevsky et al [5]. Various Pre-trained CNN models are available such as VGG16, VGG19, ResNet50, InceptionV3 which can help in image-related classification tasks. This work also adopts the idea of transfer learning [6] which helps to detect the abnormalities in crop images. Supervised machine learning techniques are also employed to predict crop damage. These models can definitely assist farmers in the early prediction of diseases and damages in crops or plants automatically.

The idea behind presenting a paper like this was to come up with a study that gathers all state-of-the-art approaches on the multimodal systems for precision agriculture use. When looking at research papers and talking to researchers in the area of precision agriculture, we learned that there is a slight disconnect between what researchers believe state-of-the-art and what actually is state-of-the-art. Our biggest contribution for this paper is not an approach, but a study of approaches with an aim to help fellow researchers leverage our findings to push the state-of-the-art further. The following are the notable contributions of this study:
\begin{itemize}

\item To the best of our knowledge, this is one of the initial attempts to present beginner-friendly extensive work related to precision agriculture systems.

\item Well-defined figures, including flowcharts and circuit diagrams, are additionally included to better understand the proposed IoT systems.

\item We successfully leverage deep learning and machine learning methods that help oversee new researchers to follow the examination work from traditional and non-conventional approaches.

\item A bird's eye view on current techniques are provided that will assist readers further push the state-of-the-art systems.  
\end{itemize}

The remaining sections are compiled as follows: Related works where we discuss the developments made in the smart precision-based farming system. The Methodology section contains the methods adopted in this project. The result segment provides the evaluation of machine and deep learning models using different metrics and last is the conclusion section.

\section{Related works}

In this section, an bird's eye view of studies and research work is provided that has been conducted in precision-based agriculture from all over the world. %Introducing IoT technologies in agriculture will help to change the traditional farming methods.  It gives us the solution to many farming issues like irrigation, land suitability, crop production, and crop disease control.%

Soil is a significant part in the field of agriculture because overall yield development is dependent on the soil so its sampling is needed to make further critical decisions. The primary goal of the soil examination is to check the strength of a land i.e. whether a field is nutrition deficient or not. Depending upon the soil conditions and weather, soil tests can be done accordingly [7]. The factors that can be determined by analyzing soil are soil types, irrigation, moisture levels, etc. These elements provide an overview regarding the synthetic, physical and natural status of the soil. Currently, there are various toolkits and sensors available to check the soil quality. These toolkits help to monitor the different soil behaviour, for example, water-holding limit, strength, and so forth. A testing kit is developed by Agrocares [8] that may test up to 100 soil samples per day. Additionally, distinguishing polluted soil by utilizing IoT innovations can further shields the field from overfertilization and yield loss [9].

Drought is also one of the serious issues that farmers face. Remote sensing is being used to deal with these problems which helps to analyze water content in the soil . Soil Moisture and Ocean Salinity (SMOS) satellites were launched in 2009. The author in [10] utilizes SMOS L2 to compute the soil water deficit index (SWDI). In [11], the researchers used a moderate resolution imaging spectroradiometer (MODIS) to gather information about the land degradation risk. In [12], sensor and a vision-based self-governing robot named Agribot has been proposed that can help in planting seeds. To determine the seed flow rates a sensor which is equipped with LEDs is used [13]. The sign data is utilized to gauge the stream rate.

About 3\% of freshwater is accessible on the earth of which around 66\% is frozen as icy masses and polar ice covers [14]. In brief, the whole world relies on 0.5\% of total water. Various traditional methods are currently employed by farmers like drip irrigation and sprinkler which are ineffective and crop production is badly affected by them. Crop efficiency stress index (CWSI) based smart irrigation is proposed in [15]. Every sensor is associated to gather the estimations and it further sends the information to the advanced calculation centre to analyze farm data through various intelligent software applications. %VRI (variable-rate irrigation) optimized by crop metrics improves water use efficiency.%

Various IoT-based fertilizers approaches are being used to estimate the nutrient requirements like Normalized Difference Vegetation Index (NDVI). It utilizes satellite pictures to check crop status [16] and is dependent on the impression of apparent and close infrared light from vegetation used to decide crop wellbeing. Advances like GPS exactness [17], variable rate innovation (VRT) [18], are added to the intelligent system. There are other sensors that also help to gather data regarding plant health and pest situations like IoT-based automated traps [19]. It is used to count and characterize insect types. An IoT-based automated robot can locate and deal with pest problems.

Machine and Deep learning methods have also been leveraged by researchers for the prediction tasks like yield prediction, object classification, multimodal frameworks [20], [32]. Arun Kumar et al. [21] utilized ANN for regression analysis to anticipate crop yields dependent on yield efficiency. Authors in [22] used time series forecasting methods to analyze and predict the weather patterns. Factors such as environmental, soil, weather, and abiotic features are adopted in [23] to classify and foresee the groundnut yield using Random Forest, SVM, and KNN.  Authors in [24] accentuate the utilization of a minimal expense UAV framework with a vision-based arrangement for the isolation of fundamental harvests from weed. It targets lessening the spread of herbicides and pesticides on crops to safeguard their quality. Strategies like harvest location are also utilized where the picture of the yield is specifically masked out of the background that incorporates soil and different items. %At that point, the vegetation cover goes through the process of feature extraction, essentially utilizing 2 techniques: Objects-based or Keypoint-based. Moreover, further geometric highlights are removed by conveying the spatial connection between the harvests. Contrasts in distance and azimuths between the inquiry objects and adjoining key points are created and utilized as a contribution for the classification. At last, the generated feature vectors are used as input to different Machine Learning algorithms for multi-classification.% 

\section{Methodology}

In this work, we implemented the smart irrigation system, smart fertilizer dose recommendation system, crop disease detection system and crop damage prediction system. This section is divided into its various subsections for detailed discussion.

\subsection{Irrigation System}
In smart irrigation method, soil moisture sensor gathers the information of the moisture content present in soil and sends it to the Node MCU. It checks the condition whether to turn on or off the water pump by a program designed in Arduino IDE. Soil moisture sensor gives the values in the range of  ADC that varies from (0 to 1023) using, 
\begin{equation}
Analog Output = ADC Value / 1023
\end {equation}
\begin{equation}
Moisture  = 100 - (Analog output * 100)
\end {equation}
The proposed flow of events is given in figure 1. Generally, moisture below 50\% is considered to be dry conditions which suggests the need to irrigate the field and above 50\% to be wet. If the value of the moisture content received from soil moisture sensor is less than the threshold value of moisture content required for the crop, it means that there is a need to irrigate the field. Node MCU sends a signal to the relay module to turn on the water pump which results in an increase in the water levels in the soil until it reaches the threshold value of moisture required for the crop. Once it reaches the threshold value, the relay module will automatically turn off the water pump. Node MCU also sends data to the firebase database. Firebase provides an online database to store data that can be extracted by the web portal. It can easily monitor the information about the moisture content of the field. The web portal also shows the information whether the water pump is on or off. We can also manually turned on and off the water pump through the web portal as shown in figure 2.

\begin{figure}[]
\centerline{\includegraphics[width=8
cm, height=5cm]{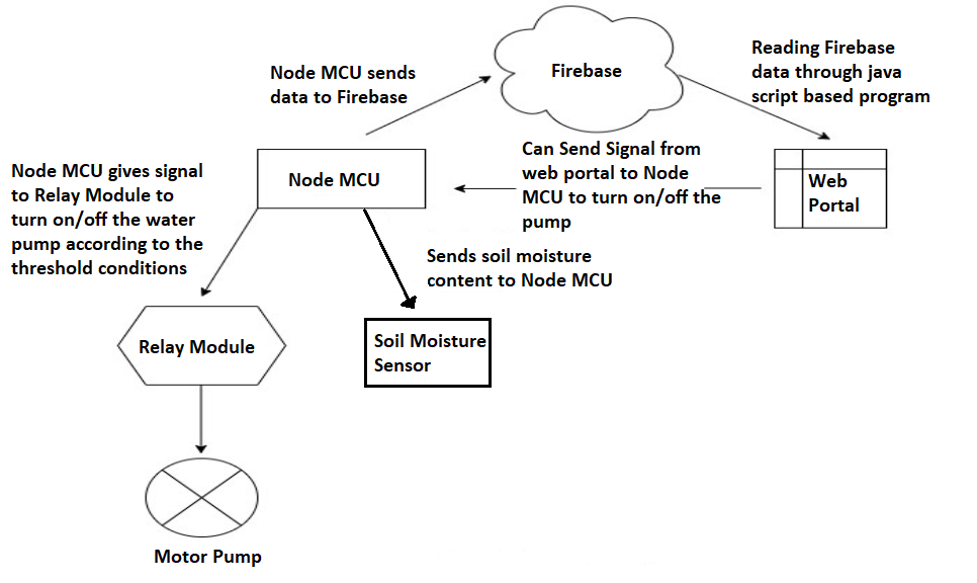}}
\caption{Proposed flow of events smart irrigation system}
\label{fig}
\end{figure}

\begin{figure}[]
\centerline{\includegraphics[width=8
cm, height=4cm]{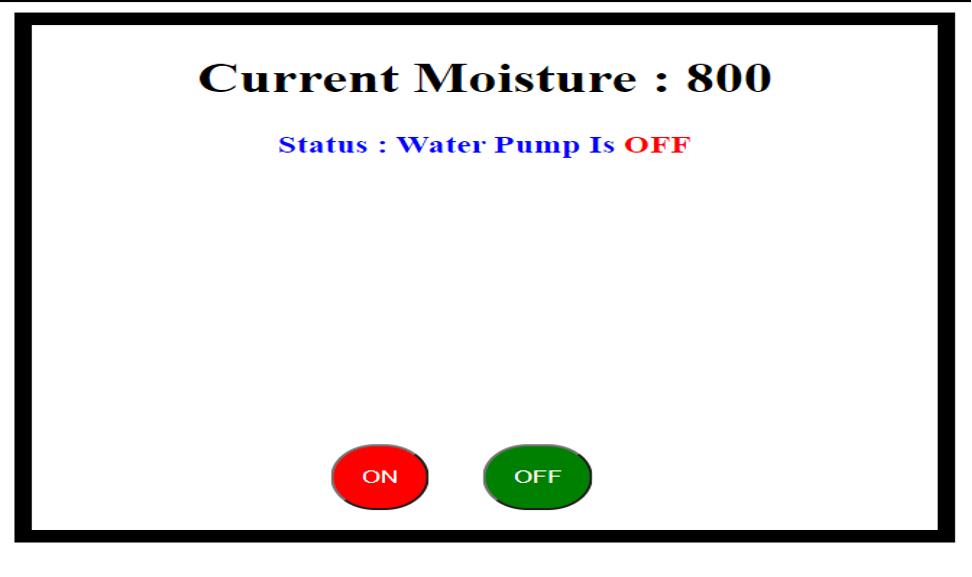}}
\caption{Web Portal for Irrigation system}
\label{fig}
\end{figure}

The circuit diagram for smart irrigation system is given in Figure 3. The SIG pin of moisture sensor provides the analog signals that is attached to A0 pin of the Node MCU and reads the analog values. VCC pin of the soil moisture sensor is connected to 3v3 pin of Node MCU for the power supply. The ground pins of both are connected with each other to use it as (0V) reference to all other electronic parts. Motor pump is attached to the NodeMCU with ground and a Vin pin is used.% for the power supply.% 
In the Relay module, the Node MCU connection with the ground pins of both the devices are connected with each other. Relay module power is connected with D1 pin of the Node MCU and provides the data from the relay module and its signal pin is connected to Vin of the Node MCU for the power supply.

\begin{figure}[htbp]
\centerline{\includegraphics[width=8
cm, height=5cm]{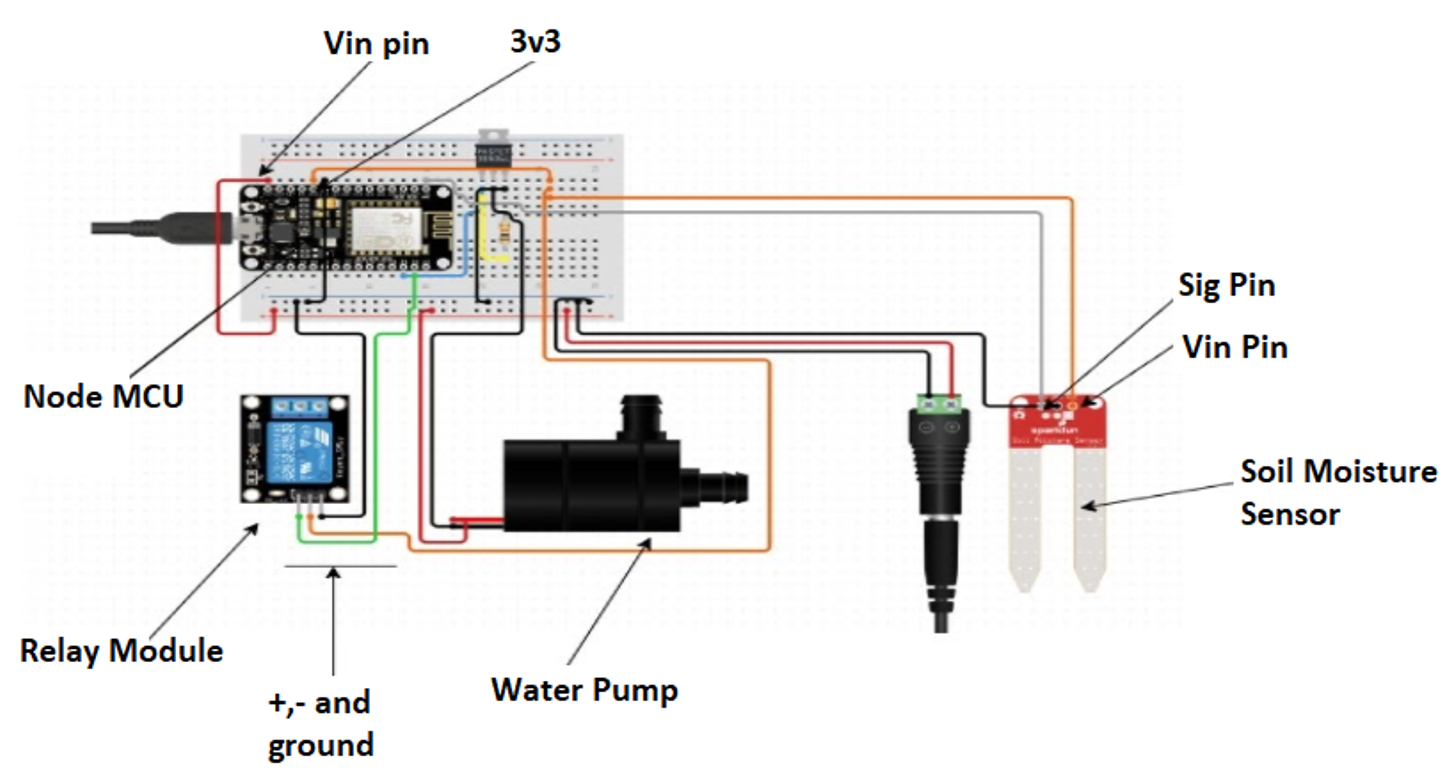}}
\caption{Circuit Diagram Smart Irrigation System}
\label{fig}
\end{figure}

\subsection{Fertilizer System}

In smart fertilizer dose recommendation system, as shown in figure 4, we first dug an n-p-k sensor into the soil to gather the data on the nitrogen, phosphorus, potassium content of the soil. The n-p-k sensor sends the n-p-k values to the NodeMCU with the help of max 485 modbus. NodeMCU sends the data to the firebase through a program design in Arduino IDE. The values can be easily accessed by a web portal given in figure 5 to perform basic calculations that helps suggesting the quantity of fertilizers for a specific crop type. For instance, the n-p-k  value of the soil given by the n-p-k sensor is (10, 5, 10 kg/ha) and the requirement is to grow wheat on this land. The n-p-k requirement of wheat is (100,20,60 kg/ha). The target is to calculate the doses of fertilizers through urea, muriate of potash (MOP) and decomposed organic phosphorus (DOP). The difference in n-p-k values is (90, 15, 50) which implies that we need 90 kg of nitrogen, 15 kg of phosphorus and 50kg of potassium.
\begin{figure}[htbp]
\centerline{\includegraphics[width=9
cm, height=5cm]{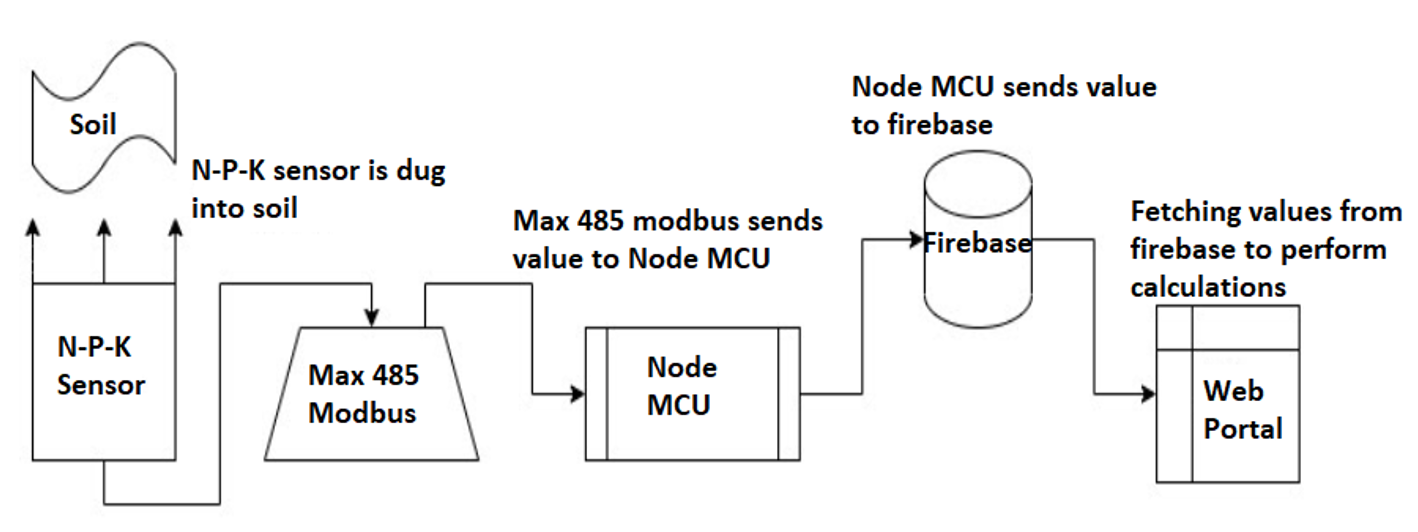}}
\caption{Proposed flow of events smart fertilizer system}
\label{fig}
\end{figure}

\begin{equation}
MOP = 60\% potassium 
\end{equation}
\begin{equation}
Urea = 46\% nitrogen 
\end{equation}
\begin{equation}
DAP = (18\% nitrogen + 46\% phosphorus)
\end{equation}
From 3, 4 and 5 we get,
\begin{enumerate}
\item 100 kg MOP  → 60kg Potassium, 

\item 50kg potassium → (100/60) * 50 of MOP = 8.3 kg of MOP
\item 100 kg DOP → 18kg of nitrogen and 46kg of phosphorus

\item 15kg phosphorus → (100/46) * 15 of DOP = 32.6kg of DOP

\item DOP also has Nitrogen, 32.6kg of DOP → (18/100) * 32.6 nitrogen = 5.86kg of nitrogen

\item Now, we require (90 - 5.86) kg of nitrogen = 84kg of nitrogen

\item 100kg of urea → 46kg of Nitrogen

\item 84kg of nitrogen → (100/46) * 84kg  of Urea = 182.6 kg of urea
\end{enumerate}
For better wheat productivity we need 8.3kg of MOP, 32.6kg of DOP and 182.6 kg of urea on this type of soil.
\begin{figure}[htbp]
\centerline{\includegraphics[width=8
cm, height=6cm]{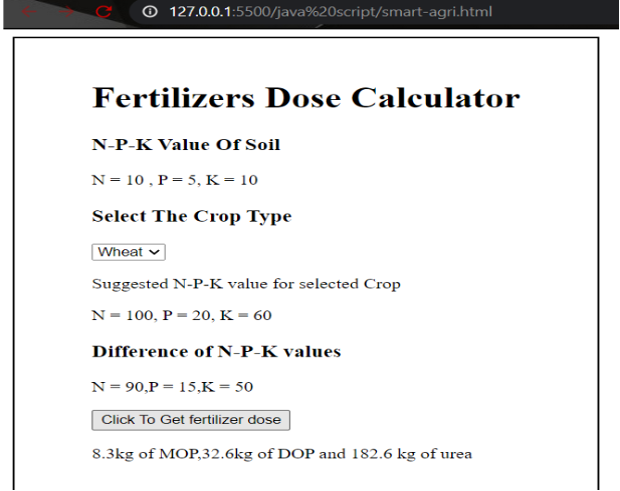}}
\caption{Web Portal for Fertilizer Recommendation}
\label{fig}
\end{figure}

The circuit diagram is presented in Figure 6. R0 and d1 pins of Modbus are connected to d2 and d3 of node MCU which sends the data collected by the n-p-k sensor to node MCU via d2 and d3 pins. The brown wire is VCC which needs a 9v-24v power supply. The Ground pin of NodeMCU is connected to the Ground pin of Modbus (black Wire). Ground pin maintains a reference level to all the other IOT connections (i.e 0v). Blue wire which is the B pin is connected to the B pin of MAX485 and Yellow wire is pin A connected to A pin of MAX485 which sends data from n-p-k to modbus.
\begin{figure}[htbp]
\centerline{\includegraphics[width=8
cm, height=4.5cm]{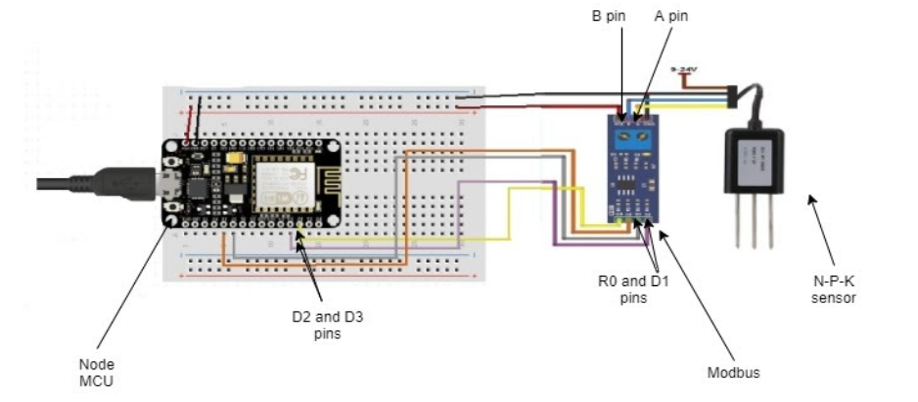}}
\caption{Circuit Diagram smart fertilizer system}
\label{fig}
\end{figure}

\subsection{Crop Disease Detection System}

In crop disease detection, we employed plant village dataset [25] containing approximately 54,000 images distributed in 38 classes. In this work, images of only six crops namely, potatoes, tomatoes, corn, peach, apple, grapes are utilized for classification. The dataset distibution of crop types used in this work is given in Table I. The images originally are of size 256*256 and are cropped to 64*64. An image augmentation pipeline using Image Data Generator is used to automatically augment the original images incorporating methods like flip, rotate, zoom in, zoom out, blur, rotate, etc. The augmentation is only performed for the training set to increase the accuracy and to avoid data leakage problems. Prior augmentation, the images were divided using the train test split method. We performed a standard 60:20:20 split in which 60 percent of total images belong to the train set and 20 percent to the test set. The validation set is also created to help train the model effectively which constitutes 20 percent of the total images.

\begin{table}[htbp]
\caption{Distribution of images}
\begin{center}
\begin{tabular}{|c|c|c|c|c|c|c|}
\hline
\textbf{Crop Type} & \textbf{Classes}& \textbf{Images}& \textbf{Train}& \textbf{Validation}& \textbf{Test}\\
\hline
Tomato& 10&18160 &10896&3632&3632  \\
\hline
Potato& 3&2152 &1291&431&430  \\
\hline
Apple& 4&3171 &1902&635&634  \\
\hline
Peach& 2&2657 &1594&532&531  \\
\hline
Grapes& 4&4062 &2437&813&812   \\
\hline
Corn& 4&3852 &2371&771&770  \\
\hline
\end{tabular}

\end{center}
\end{table}

The concept of transfer learning is used in this project to transfer the weight of the already trained models i.e. Pre-Trained CNN (Convolutional Neural Network) models [6]. It helps in reducing the computational power and speeds up the performance which shows indications of quicker and improved outcomes. In this study, we utilize only three models namely ResNet50, VGG16, and DenseNet121 [26], [27], [28] for comparative analysis. Our proposed pipeline for preparing the model contains three areas:
\begin{itemize}
    \item Feature Generation - Separating the main features by calibrating the models through preparing the three State of the Art (SOTA) Pre-trained CNN structures. The extracted features after this phase act as an input to the modification phase.
    \item Modification - Incorporated a concat layer which is a concatenation of features of three layers namely, MaxPooling2D, AveragePooling2D, flatten layer. In addition, a dropout layer with a dropout rate of 0.5 has additionally been fused.
    \item Output - The output generated from the modification phase goes to the dense layer. The activation function used is sigmoid. For compilation of model , adam optimizer is used with a learning rate of 0.0002. The batch size is 32 and the model is trained for 30 epochs.
\end{itemize}

\subsection{Crop Damage Prediction System}

In crop damage prediction, the task is to use and employ feature engineering concepts with machine learning algorithms to predict the category of crop damage. We used 'machine learning in agriculture' dataset from the Analytics Vidhya website [33]. This dataset has 88858 rows and 10 columns which is divided into 75:25 using train test split. The description of the features are given in Figure 7.

\begin{figure}[htbp]
\centerline{\includegraphics[width=8
cm, height=3.5cm]{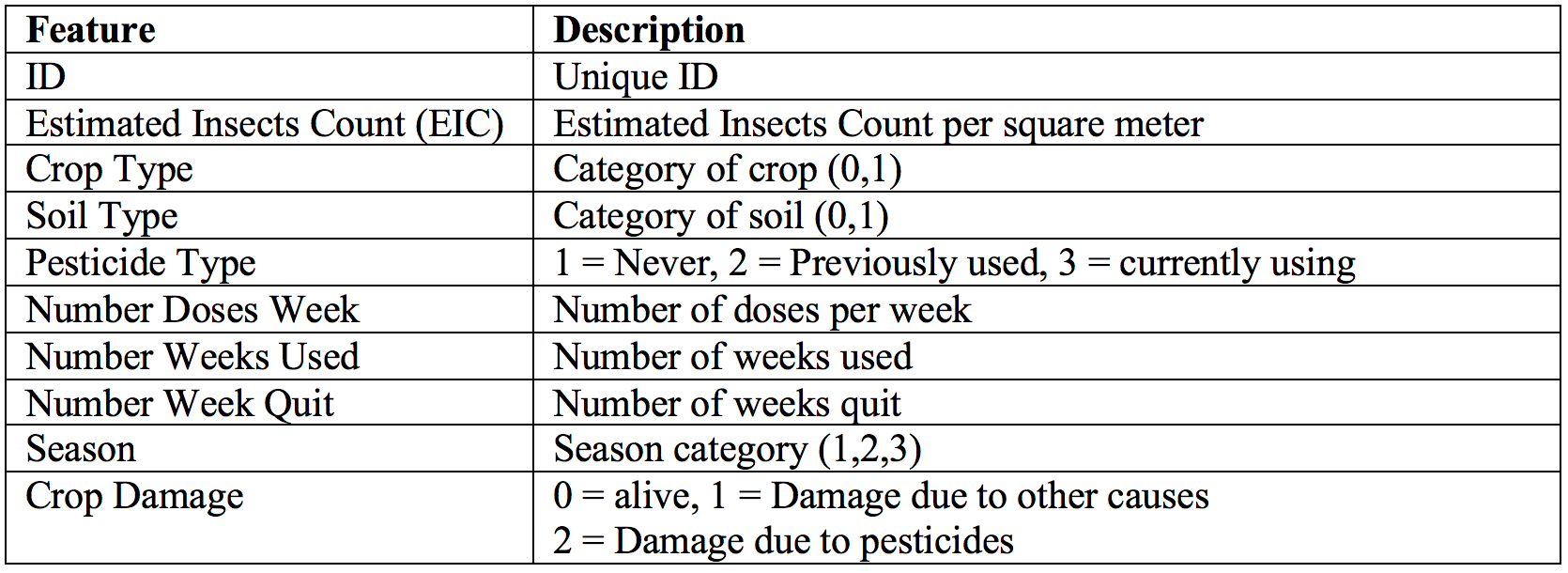}}
\caption{Dataset Description}
\label{fig}
\end{figure}

It is important to clean and visualize data before implementing machine learning algorithms. This dataset initially contains null values replaced by -999 using fillna method. It also contains columns like Estimated insects counts as shown in Figure 8 with some time series related pattern that suggests to generate features containing different lag values using shift and rolling method. A window size of 5 is used to extract the mean of five observations with a lag of 1 and 2.

\begin{figure}[htbp]
\centerline{\includegraphics[width=8
cm, height=3cm]{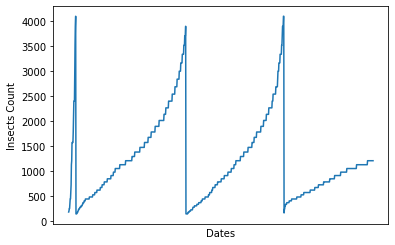}}
\caption{Estimated Insects Count}
\label{fig}
\end{figure}

We used five machine learning algorithms, namely, Random Forest, LGBM, KNN, Decision Tree and XGBoost [29], [30], [31], [9] for classification. Decision Trees is a simple supervised learning algorithm built on the concepts of trees. The tree has two components namely leaves and nodes. The leaves are the choices or the ultimate results whereas the nodes  represent the points where the data is divided according to a fixed parameter for further classification. Random Forest is an ensemble-based supervised machine learning algorithm. The bagging method is used to train the ensemble of decision trees which help to build the forest. In simple terms, Random Forest assembles numerous decision trees and combines them to get a more exact and stable expectation. K-nearest neighbors (KNN) depends on the possibility of the likeness of similarities. It predicts the values of new points dependent on how closely the new information is identified with the upsides of the given training set. The KNN doesn't make any assumptions identified with the preparation information and it for the most part inputs all the data values for training. 

Extreme gradient boosting (XGBoost) and light gradient boosting (LGBM) are based on the concept of gradient boosting. This technique leverage weak models utilizing ensemble methods that can develop new models to decrease the loss function estimated by gradient descent. However, there is a subtle difference these two models in terms of split. For XGBoost, histogram based filters are used while for LGBM gradient-based one-side sampling (GSOS) method is used to distinguish the best split. It has been observed that  histogram based filters technique is computationally costly compared with GSOS which recommends that LGBM is more effective than XGBoost.

\subsection{Metrics}
In this research, we primarily perform two prediction tasks to classify crop damages and disease detection. Since both fall under the category of classification,  the models are evaluated by metrics namely, precision, recall, f1 score and accuracy. 
\begin{equation}
Precision = \frac{T_{p}}{T_{p}+F_{p}}
\end{equation}
\begin{equation}
\\Recall = \frac{T_{p}}{T_{p}+F_{n}}
\end{equation}
\begin{equation}
Accuracy = \frac{T_{p}+T_{n}}{T_{p}+T_{n} + F_{p}+F_{n}}
\end{equation}
\begin{equation}
F1score = 2.\frac{Precision.Recall}{Precision+Recall}
\end{equation}
where $T_p$, $T_n$, $F_p$, $F_n$ represents True positive, True negative, False positive, False negative.

\section{Results}
Table II shows the evaluation of metrics for crop disease detection. We performed the disease classification on six crops, tomato, potato, apple, peach, grapes and xorn using three Pre-trained CNN models, VGG16, ResNet50, Densenet121. All models performed well and there is a slight difference between the accuracy of the models. For tomato, corn and peach, Densenet121 outperforms other models in terms of all evaluation metrics. VGG16 also performed state of the art results for potato, apple and grapes. However, it marginally underperformed in classification for peach and tomato. The results forecasted from ResNet50 manifests that aside from class apple it failed to outperform other models. For better interpretation and visualization, bar plot is pictured in figure 9 showing precision of models on given classes.

%ResNet50 completely outperformed VGG16 and DenseNet121 in terms of all evaluation metrics. The results forecasted from VGG16 manifests that the model needed more number of epochs for training. Models which are complex will perform better in low number of epoch training as well. For better interpretation and visualization, bar plot is also pictured in Figure 9.
\begin{table}[htbp]
\caption{Evaluation of Metrics for crop disease detection}
\begin{center}
\begin{tabular}{|c|c|c|c|c|c|c|}
\hline
\textbf{Crop Type} & \textbf{Model}& \textbf{Prec}& \textbf{Rec}& \textbf{F1}
& \textbf{Acc.}\\
\hline
& VGG16& 0.93&0.94 &0.94&0.945  \\
Tomato& ResNet50& \textbf{0.96}&0.96 &0.96&0.965\\
& Densenet121& \textbf{0.96}&\textbf{0.97} &\textbf{0.97}&\textbf{0.973}\\
\hline
& VGG16& \textbf{0.94}&0.96 &\textbf{0.95}&\textbf{0.981} \\
Potato& ResNet50& 0.89&\textbf{0.97} &0.92&0.960\\
& Densenet121& 0.87&0.96 &0.90&0.955\\
\hline
& VGG16&\textbf{0.99}&\textbf{0.98} &\textbf{0.98}&\textbf{0.981} \\
Apple& ResNet50& 0.98&0.96 &0.97&0.970\\
& Densenet121& 0.97&0.95 &0.96&0.970\\
\hline
& VGG16& 0.97&0.98 &0.97&0.986 \\
Peach& ResNet50& \textbf{1.00}&\textbf{0.99} &\textbf{0.99}&\textbf{0.996}\\
& Densenet121& \textbf{1.00}&\textbf{0.99} &\textbf{0.99}&\textbf{0.996}\\
\hline
& VGG16& \textbf{0.97}&\textbf{0.97} &\textbf{0.97}&\textbf{0.969} \\
Grape& ResNet50& 0.96&0.96 &0.96&0.948\\
& Densenet121& \textbf{0.97}&0.96 &0.96&0.963\\
\hline
& VGG16& 0.93&0.93 &0.93&0.946 \\
Corn& ResNet50& 0.93&0.93 &0.93&0.943\\
& Densenet121& \textbf{0.95}&\textbf{0.95} &\textbf{0.95}&\textbf{0.965}\\
\hline
\end{tabular}

\end{center}
\end{table}

\begin{figure}[htbp]
\centerline{\includegraphics[width=8
cm, height=5cm]{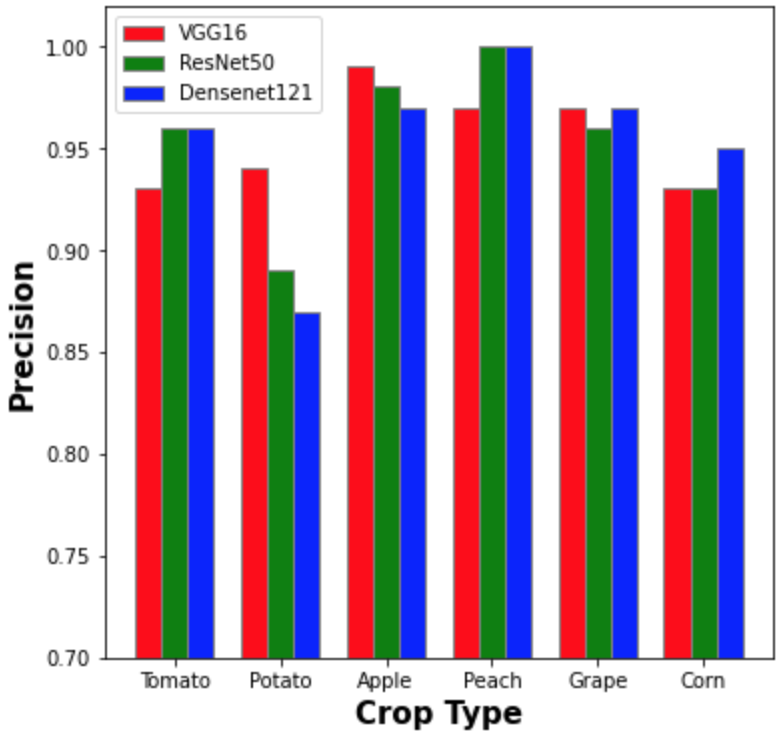}}
\caption{Bar Plot showing precision for Crop Disease Detection}
\label{fig}
\end{figure}

Analysis of evaluation metrics are also discussed for crop damage prediction. As we can see from Table  III, the highest accuracy achieved is 94\% by LGBM and it completely beats other algorithms. Random Forest and XGBoost have very subtle difference in their predictions. KNN is the worst performing algorithm forestalling less than 5 percent precision for minority classes. The differences in the results of the various classes suggest that the algorithms failed to manage the issue of class imbalance. In general, tree based classifiers outperformed the standard machine learning algorithm. A bar plot is pictured in Figure 10 shows the precision of machine learning algorithms used for crop damage prediction on various sub classes.
\begin{table}[htbp]
\caption{Evaluation of Metrics for crop damage prediction}
\begin{center}
\begin{tabular}{|c|c|c|c|c|c|}
\hline
\textbf{Model } & \textbf{Class}& \textbf{Prec}& \textbf{Rec}& \textbf{F1} & \textbf{Acc.}\\
\hline
& 0& 0.95&\textbf{0.99} &0.97&\\
RF& 1& 0.78&0.72 &0.75&0.93\\
& 2& 0.57&0.10 &0.16&\\
\hline
& 0& \textbf{0.97}&\textbf{0.99} &\textbf{0.98}& \\
LGBM& 1& \textbf{0.82}&\textbf{0.78} &\textbf{0.80}&\textbf{0.94}\\
& 2& 0.44&0.20 &\textbf{0.28}&\\
\hline
& 0& 0.96&0.91 &0.93& \\
DT& 1& 0.50&0.59 &0.54&0.85\\
& 2& 0.19&\textbf{0.32} &0.24&\\
\hline
& 0& 0.95&0.98 &0.96& \\
XGB& 1& 0.72&0.70 &0.72&0.92\\
& 2& \textbf{0.60}&0.06 &0.11&\\
\hline
& 0& 0.85&0.97 &0.90& \\
KNN& 1& 0.23&0.06 &0.09&0.84\\
& 2& 0.05&0.01 &0.01&\\
\hline
\end{tabular}

\end{center}
\end{table}

\begin{figure}[htbp]
\centerline{\includegraphics[width=8
cm, height=5cm]{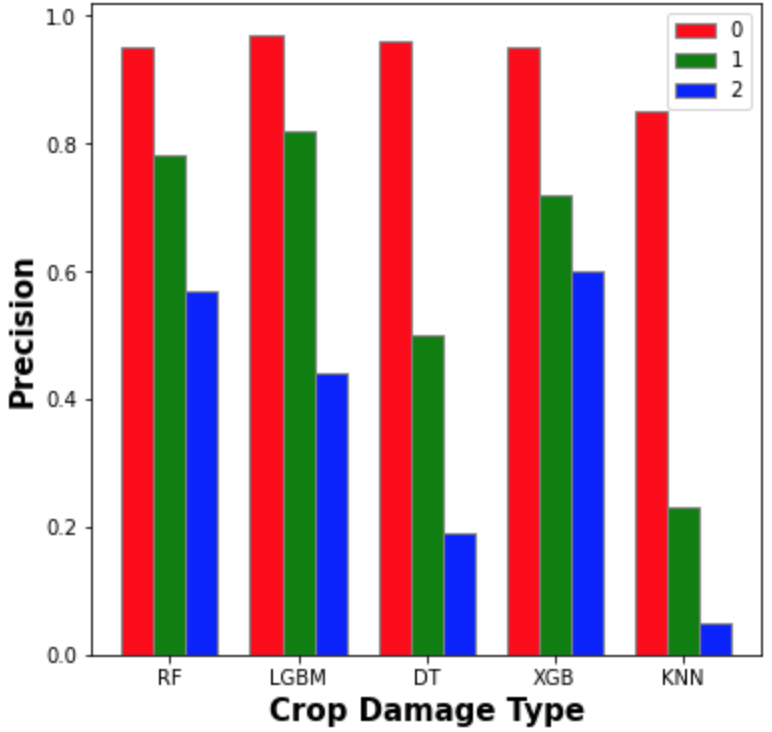}}
\caption{Bar Plot showing Precision for Crop Damage Prediction}
\label{fig}
\end{figure}

\section{Conclusion}
A multimodal precision farming system is implemented in this project which consists of intelligent fertilizer, irrigation, crop disease and damage prediction which help reduce the efforts and labors in the agriculture sector. In this work, we intend to provide the methodology with circuit diagram, flowchart and theoretical aspects for IoT systems. Our work is well-organized, easy to read, contains well-defined figures and diagrams. This would definitely help new researchers to get a better understanding of the problem statement and further push the state of the art systems. Subsequently, we leverage multimodal approach to train an image classification and machine learning models related to crop disease detection and damages. In multiclass image classification, we employed three Pre-Trained CNN architectures namely, ResNet50, VGG16, DenseNet121. In crop damage prediction, LightGBM beats XGBoost and Random Forest by a slight margin. However, the aformentioned models aren't able to produce quality results for minority classes. Oversampling techniques like SMOTE and ADASYN are required to curb the dominance of majority class. Hyperparameter optimization techniques such as evolutionary algorithms are missed in this research due to time constraints and less computational power. It will be a part of the future work. 

\end{document}